\documentclass[runningheads]{llncs}
\usepackage[T1]{fontenc}
\usepackage{graphicx}
\usepackage{bbold}
\usepackage{url}
\usepackage{cite}
\usepackage{todonotes}
\usepackage{booktabs}

\usepackage{amsfonts}

\begin{document}
\title{Expert-aware uncertainty estimation for quality control of neural-based blood typing}
%
%
%
%

%

\author{Ekaterina Zaychenkova\inst{1,2}, Dmitrii Iarchuk\inst{1,2}, Sergey Korchagin\inst{1,2},\newline Alexey Zaitsev\inst{3}, Egor Ershov\inst{1, 4}}

\institute{Institute for Information Transmission Problems,
\and
Moscow Institute of Physics and Technology,
\and
The Skolkovo Institute of Science and Technology
\and
AIRI\\
\email{zaichenova.ee@phystech.edu, iarchukdima@gmail.com, korchagin.sergey.97@gmail.com,
a.zaytsev@skoltech.ru, e.i.ershov@gmail.com}}

\authorrunning{Ekaterina Zaychenkova, Dmitrii Iarchuk, Sergey Korchagin}
\titlerunning{Expert-aware uncertainty estimation}

\maketitle              
\begin{abstract}
In medical diagnostics, accurate uncertainty estimation for neural-based models is essential for complementing second-opinion systems.
Despite neural network ensembles' proficiency in this problem, a gap persists between actual uncertainties and predicted estimates. 
A major difficulty here is the lack of labels on the hardness of examples: a typical dataset includes only ground truth target labels, making the uncertainty estimation problem almost unsupervised. 

Our novel approach narrows this gap by integrating expert assessments of case complexity into the neural network's learning process, utilizing both definitive target labels and supplementary complexity ratings. 
We validate our methodology for blood typing, leveraging a new dataset "BloodyWell" unique in augmenting labeled reaction images with complexity scores from six medical specialists.
Experiments demonstrate enhancement of our approach in uncertainty prediction, achieving a 2.5-fold improvement with expert labels and a 35\% increase in performance with estimates of neural-based expert consensus.

\end{abstract}

\section{Introduction}~\label{sec:introduction}

\begin{figure}
    \includegraphics[width=\textwidth]{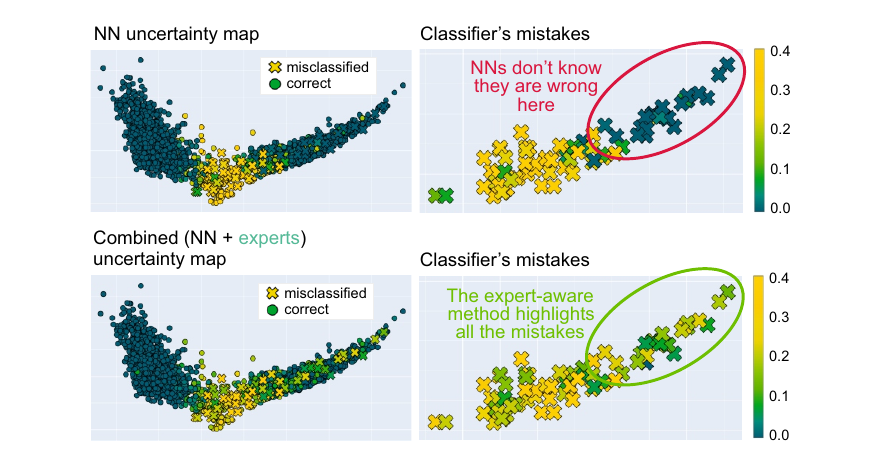}
    \caption{The feature space of a classification neural network with uncertainty estimation by a classical approach and by an expert-aware approach proposed by us. 
    } \label{fig:frontal}
\end{figure}

Second-opinion systems based on AI are becoming more and more widespread in medicine~\cite{briganti2020artificial, kurz2022uncertainty}. 
Many studies show the faster diagnostic capabilities of AI~\cite{eng2021artificial} and compatible with health-care professional diagnostics quality~\cite{liu2019comparison}.
By providing auxiliary decisions, these systems act as a safeguard, potentially preventing professional oversight and guiding medical experts toward more accurate diagnoses~\cite{punzi2024ai}.

To guarantee reliable integration of AI-based second opinions into the medical diagnostic process, one should ensure that the system not only delivers accurate results but also provides a measure of confidence in those results~\cite{abdullah2022review, huang2023review}. 
This capability is crucial as it enables laboratory technicians to prioritize samples based on the likelihood of error, thereby optimizing workflow and focusing on areas needing closer examination.
This challenge falls within the domain of neural network uncertainty estimation (UE), a process that quantifies the confidence of the model in its predictions~\cite{kabir2018neural}.

Despite the fact that there are numerous methods for UE~\cite{gawlikowski2023survey}, there is a gap between true uncertainty values and their estimation~\cite{kostenok2023uncertainty}. 
Since true uncertainties are unknown, the problem belongs to unsupervised and, thus, difficult.

One possible solution to this problem is to use the markup of experts as an alternative uncertainty estimation~\cite{ayhan2020expert}.
Human experts' extensive experience allows them to better assess data uncertainty, i.e., aleatory uncertainty (AU).
Simultaneously, relying solely on expert uncertainty is insufficient for assessing the uncertainty of a model, as individuals and neural networks frequently consider various factors.

In this work, we, for the first time, propose to advance the problem of UE for neural networks by training a model that looks at both true labels and uncertainty estimation provided by experts.
Our method naturally combines these two sources of information and provides superior uncertainty estimates for neural networks, see Figure \ref{fig:frontal}.

\textbf{Contribution.} We propose a new expert-aware uncertainty quantification (EAUQ) method, which provide one and half times more precise uncertainty estimation.
The fundamental concept behind this method is to determine full uncertainty by combining data-based (aleatory) uncertainty estimated using experts and model-based (epistemic) uncertainty from NNs ensemble. 
To validate the proposed method we publish a novel large BloodyWell dataset on blood typing. 
It comprise 3139 samples, each with the actual blood type from medical records and assessments from six experts.
The collected data will be useful for developing solutions to the problem of uncertainty estimation and blood typing.
\section{Literature review}\label{sec:lit_review}

Given the importance of UE in neural networks, a substantial body of research has emerged in this area~\cite{gawlikowski2023survey}.
The majority of methods belong to two categories:
\begin{enumerate}
    \item \textit{Single network deterministic methods} give the prediction based on one single forward pass within a deterministic network.
    \item \textit{Bayesian methods} cover all kinds of stochastic DNNs and posterior analysis of several different deterministic networks at inference.
\end{enumerate}
In this paper, we decided to focus on the first two categories since they correspond closely with our case, where we have both GT from medical cards and expert assessments.

The primary benefit of the first category lies in the straightforward nature of the problem statement: we can choose to view network confidences as a way to quantify uncertainty~\cite{sensoy2018evidential, malinin2018predictive} or to train a distinct network for this purpose~\cite{raghu2019direct, ramalho2020density}.

A more principled way to UE is the Bayesian treatment.
It provides a probabilistic interpretation of the model's outputs, replacing point estimates with distributions or confidence intervals.
So, the variance of a predictive distribution reflects the uncertainty of the prediction~\cite{gruber2023sources}.
While being introduced for neural networks quite early~ \cite{denker1990transforming}, this nice idea comes with its own challenges. 
Exact distributions are analytically intractable, so one relies on approximations such as variational inference \cite{hinton1993keeping} or Monte Carlo dropout \cite{gal2016dropout}. 

A successful approximation of Bayesian techniques involves training a diverse ensemble of models to treat discrepancies in their outputs as uncertainty estimates~\cite{lakshminarayanan2017simple}.
Neural networks are made as variable as possible so that they represent different feature spaces.
The diversity comes from using different neural network architectures~\cite{herron2020ensembles}, training with different data subsamples~\cite{bishop2006pattern}, or using different augmentations.
While being rather accurate, this approach has limited capacity to match the true distribution~\cite{wilson2020bayesian} and requires runs of many models instead of one.

Due to the nature of uncertainty, it is impossible to match its ground truth values without seeing examples of uncertainties during training.
A reasonable practice to overcome this issue is to use the diversity of answers by several experts as the ground truth.
For example, in the work~\cite{raghu2019direct} the authors predicted the answers of experts as an uncertainty, and in the work~\cite{campos2023evolving} authors used feedback from expert uncertainty to guide a neural network evolution.
Expert assessment is also used to validate the resulting uncertainty~\cite{ayhan2020expert}.
In the work~\cite{wilder2020learning}, authors propose a hybrid system where the final classification is chosen as the best suitable value from the NN answer and expert output.
Relying solely on expert labels has limitations due to scarcity, and neural networks are better at estimating uncertainty compared to traditional methods.

While the existing landscape of research on UE is wide,
methods that take into account uncertainty from experts are rarely found, and even fewer results are available in cases where both expert labels and ground truth labels coexist (which is also limits applicability). 
So, it is desirable to develop a UE for a neural network that can naturally incorporate both sources of labels. 
\section{Expert-aware uncertainty estimation} \label{sec:methods}


\paragraph{Uncertainty decomposition.}
Our goal is to estimate the uncertainty of a binary classification model's response. 
The estimate should correlate with the error probability.
For an object $x$ and a model $p(x)$, uncertainty can be decomposed to two types: aleatoric and epistemic: $\mathrm{UQ}(x) = \mathrm{UQ}_{\mathfrak{a}}(x) + \mathrm{UQ}_{\mathfrak{e}}(x)$, where the AU $\mathrm{UQ}_{\mathfrak{a}}(x)$ corresponds to the noise in data, and the epistemic one $\mathrm{UQ}_{\mathfrak{e}}(x)$ corresponds to the lack of the model's knowledge~\cite{gruber2023sources}.
So, we need to develop a method that accounts both types.

\paragraph{Ensemble uncertainty estimates.}
A natural baseline is standard deviation of the answers of models in an ensemble:
$$
    \mathrm{STD}(x) = \sqrt{\frac{1}{k} \sum_{i = 1}^k (p_i(x) - \overline{p}(x))^2},
$$
where $p_i(x)$ is the probability of a class one by $i$-th model, and $\overline{p}(x) = \frac{1}{k} \sum_{i = 1}^k p_i(x)$ is the average probability of the ensemble. 
This estimate considers prediction diversity and achieves high accuracy in regions with little training data, where models are uncoordinated.
Thus, it captures the epistemic uncertainty, while ignoring aleatoric: $\mathrm{STD}(x) \approx \mathrm{UQ}_{\mathfrak{e}}(x)$.

Medical experts' extensive problem-solving experience enhances their ability to evaluate the complexity of various cases accurately. 
Assuming that experts have encountered diverse data variations, their assessments primarily reflect aleatoric uncertainty inherent in the data rather than uncertainties associated with a classifier itself. 
This implies that expert evaluations could reliably indicate uncertainty levels applicable to both human and algorithmic classifiers. 
Moreover, experts' answers are calibrated, as they estimate the probability of an error.

\begin{figure}
    \includegraphics[width=\textwidth]{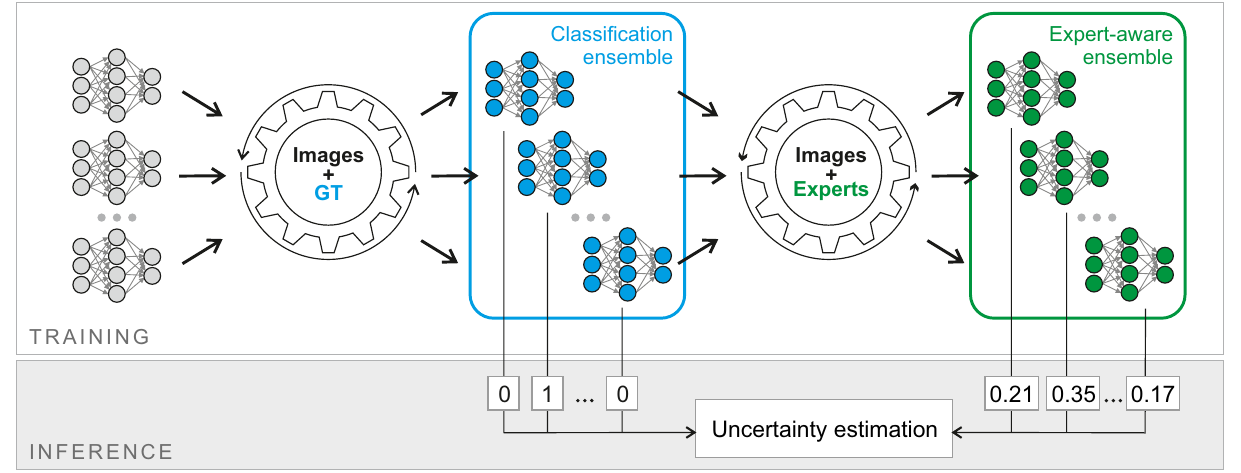}
    \caption{Proposed expert-aware uncertainty estimation training and inference pipeline.} 
    \label{fig:method}
\end{figure}
\paragraph{Experts' assessment of uncertainty.}
\label{sec:experts}

\begin{figure}  
    \centering
    \includegraphics[width=\textwidth]{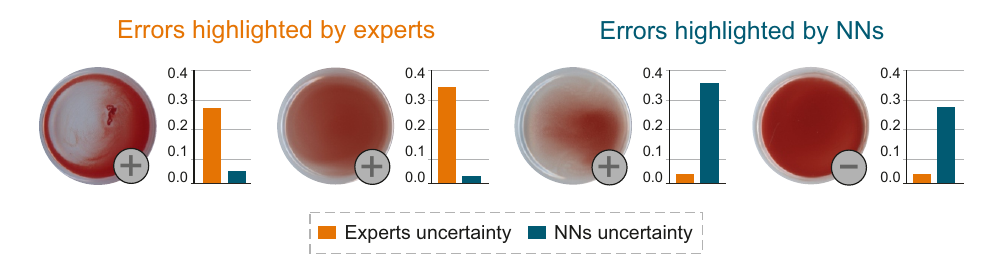}
        \caption{Examples of classification errors with uncertainty values of the experts' ensemble and the NN ensemble. Symbols plus and minus states for agglutination GT.
        } \label{fig:mistakes}
\end{figure}

We adopted a straightforward way for expert evaluation.
For binary classification assessment, $k$-th expert selected reaction probabilities $e_k$ from $0.00$ (definitely negative reaction), $0.25$ (most likely negative), $0.50$ (difficult to answer), $0.75$ (most likely positive), and $1.00$ (definitely positive).
The distance of the average experts' probability $\overline{e} = \frac1n\sum_{k=1}^{n} e_k$ to the nearest integer provides the estimate of the complexity of an example~\cite{mukhoti2023deep}:
$$
    \mathrm{MP}(x) = 1 - \max(\overline{e}(x), 1 - \overline{e}(x)).
$$
As we discussed, $\mathrm{MP}(x) \approx \mathrm{UQ}_{\mathfrak{a}}(x)$, approximating the aleatoric uncertainty.
So, $\mathrm{MP}(x)$ + $\mathrm{STD}(x)$ should help us to highlight more model errors, as shown in Figure~\ref{fig:mistakes}.
However, expert labels are unavailable during model usage, and we need to learn to imitate them.


\paragraph{Combining expert-based and ensemble uncertainty estimates.}
To enhance the model with expert estimates, we adopt the approach presented in Figure~\ref{fig:method}.

First, we trained a binary classification ensemble (CE) of 20 neural networks with GT values of agglutination outputs.  
UE is calculated via the sum of the standard deviation of produced outputs~$\mathrm{STD}$ and expert ensemble metric~$\mathrm{MP}$.
This method relies on expert annotations during inference, limiting its applicability while providing an unbiased representation of total uncertainty $\mathrm{UQ}(x)$.

We updated this uncertainty with an approximate expert surrogate via a single deterministic expert-aware network (EAN).
This NN is initialized the same as in CE and fine-tuned to predict the average response of the expert ensemble instead of a binary output using the same loss.
During additional training, the learning rate was limited to avoid overfitting.
For each net as produced uncertainty, we used the $\mathrm{MP}(x)$ of EAN output as an aleatoric approximation and added it to standard deviation~$\mathrm{STD}$ of CE.

As an extension of this idea, we estimated uncertainty via an Expert-aware ensemble (EAE), which considers all $20$ fine-tuned EAN.
For uncertainty estimation, we took aleatoric uncertainty approximation $\mathrm{MP}$ of EAE. 
To obtain full uncertainty, our method considers adding it to two different epistemic uncertainty approximations $\mathrm{STD}$ of CE and $\mathrm{STD}$ of EAE itself. 

\section{BloodyWell dataset} \label{sec:dataset}

\begin{figure}
    \includegraphics[width=\textwidth]{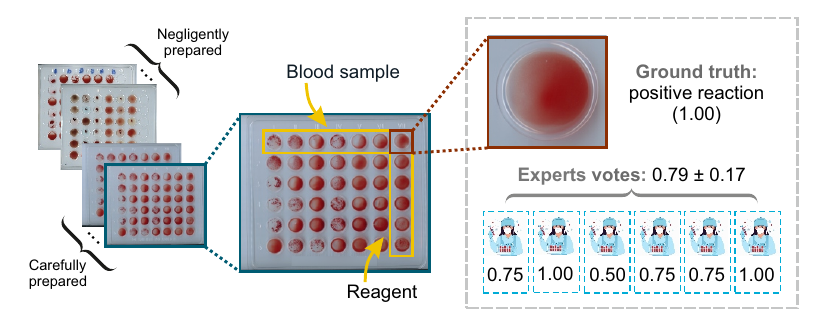}
    \caption{
    Structure of prepared serological plate with agglutination reactions. One sample of the dataset is a cut out image with markup from two sources: information about the type of agglutination obtained from the blood donor's medical record (presence or absence) and an alternative assessment of agglutination by six medical experts.
    } \label{fig:dataset}
\end{figure}

The dataset we collected in this work consists of 92 high-resolution scanned images of serological plates.
Each plate contains 42 wells with agglutination reactions: 6 rows, each corresponding to one blood sample, and 7 columns, each corresponding to one type of reagent.
In total, the dataset consists of 3139 images of non-empty wells with extensive markup (see Figure \ref{fig:dataset}).

Agglutination reactions serve to determine three blood group systems: ABO, RH, and KELL.
The commonly utilized antigens A, B, and D typically cause rapid and clear agglutination reactions. 
Responses from other reagents might be subtler, complicating visual analysis. 
Data were collected to mirror varying laboratory practices, emulating both meticulous and careless lab assistants. 


With over five years of experience in manual blood typing, all experts are qualified in serology.
Detailed description provided in~\cite{korchagin2024image}.
\section{Experimental results}\label{sec:results}

\paragraph{Experimental setup.}
The dataset was divided into training, validation, and test sets in ratios of 10/10/80\%, respectively.
We repeated the training process 20 times on different train-validation splits to provide results reliability.

As a simple and reliable classifier, we have chosen MobileNet-V3-Large architecture.
The training process spanned 800 epochs, commencing with a scheduled decreasing learning rate of 1e-4 and a 5e-5 weight decay.
Data augmentations consisted of random shifts of image borders, small Gaussian noise, and other basic augmentations. 

As baseline uncertainty estimators, we used Monte Carlo Markov Chain (MCMC)~\cite{welling2011bayesian}, MC Dropout~\cite{gal2016dropout}, and standard deviation of NNs classification ensemble (CE).
The final uncertainty for each baseline method was calculated as a $\mathrm{STD}$ over ensemble outputs.
Each MCMC ensemble comprises 10 NNs created by check-pointing the model at 15-epoch intervals during training.
MC Dropout builds an ensemble for each classification net by activating dropout layers during testing with 50 random initializations. 
The dropout probability for inference was set to 0.2, the same as for training.
CE was built from the 20 NNs described above. 
Its training process alone brings enough variety to produce different feature spaces: standard data augmentations, dropout layer usage, and different splits between train and validation sets.

For our methods, each of the 20 classification NNs was fine-tuned to predict an average expert vote, resulting in the corresponding Expert-Aware net (EAN).
Fine-tuning lasted for 40 epochs, with an exponentially decreasing coefficient of 0.99 for the 1e-5 initial learning rate, while other meta-parameters were taken from basic classification net training.

\begin{figure}[ht]
    \includegraphics[width=\textwidth]{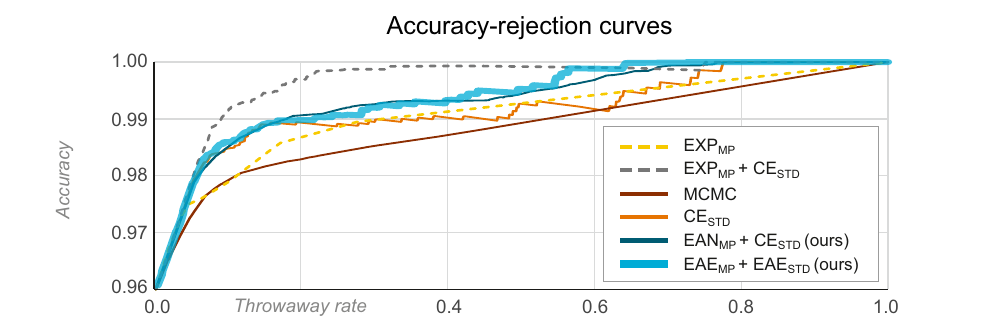}
    \caption{Accuracy-rejection curves for the main methods used. $\mathrm{EXP_{MP}}$ and $\mathrm{EXP_{MP}} + \mathrm{CE_{STD}}$ use expert uncertainty labels during inference} 
    \label{fig:rejection_curves}
\end{figure}

\paragraph{Results analysis.}
One of the most widely utilized ways to assess and compare UE techniques is rejection curves, presented in  Figure~\ref{fig:rejection_curves}. 
A more favorable method is characterized by a rapid rise in accuracy as progressively more uncertain samples are excluded.
In this study, we analyze the area above the curve (a lower area value corresponds to a better uncertainty estimation).

It is worth mentioning that the laboratory assistant may not be interested in the full trend of the rejection curve in the practical usage of second-opinion systems.
So, we also pay attention to the accuracy when excluding 10\% of test samples and to the number of samples required to be discarded to attain a 99\% accuracy level, see Table~\ref{tab:methods}.

The classification ensemble approach ($\mathrm{CE_{STD}}$) surpassed other classical methods in terms of all the criteria. 
The ensemble of medical experts ($\mathrm{EXP_{MP}}$) proved to be 1.33 times better than CE in terms of reaching a $99\%$ accuracy level having almost the same AAC.

\begin{table}[h]
    \centering
    \caption{Comparison of uncertainty estimation methods. Second column: area above the rejection curve (AAC), third: accuracy upon discarding 10\% of the most uncertain data. The last two methods use expert uncertainty estimates during inference}
    
    \begin{tabular}{lccc}
    \toprule
    \vtop{\hbox{\strut }\hbox{\strut \textbf{Method}}} & 
    \vtop{\hbox{\strut AAC, $\downarrow$,}\hbox{\strut$\times$1e-4}} & 
    \vtop{\hbox{\strut Discarding}\hbox{\strut 10\% data}\hbox{\strut accuracy, $\uparrow$, \%}} & 
    \vtop{\hbox{\strut Part of dataset}\hbox{\strut omitted to attain}\hbox{\strut 99\% accuracy, $\downarrow$, \%}}\\
    \midrule
    \textbf{$\mathrm{\textbf{CE}_{\textbf{STD}}}$} & 89 & 98.35 & 48.3 \\
    
    \textbf{MCMC} & 124 & 97.76 & 56.9 \\
    
    \textbf{MC Dropout} & 153 & 97.98 & 97.9\\
    
    \textbf{$\mathrm{\textbf{EAE}_{\textbf{MP}}}$ (ours)} & 69 & \textbf{98.57} & 24.0 \\
    
    \textbf{$\mathrm{\textbf{EAN}_{\textbf{MP}}}$  + $\mathrm{\textbf{CE}_{\textbf{STD}}}$ (ours)} & 69 & 98.42 & 23.3 \\

    \textbf{$\mathrm{\textbf{EAE}_{\textbf{MP}}}$ + $\mathrm{\textbf{CE}_{\textbf{STD}}}$ (ours)} & 68 & 98.54 & \textbf{20.4} \\

    \textbf{$\mathrm{\textbf{EAE}_{\textbf{MP}}}$ + $\mathrm{\textbf{EAE}_{\textbf{STD}}}$ (ours)} & \textbf{66} & 98.48 & 24.7\\

    \midrule
    \textbf{$\mathrm{\textbf{EXP}_{\textbf{MP}}}$} & 98 & 97.73 & 36.3 \\
    \textbf{$\mathrm{\textbf{EXP}_{\textbf{MP}}}$ + $\mathrm{\textbf{CE}_{\textbf{STD}}}$} & 36 & 99.11 & 9.4 \\
    \end{tabular}
    \label{tab:methods}
\end{table}

Combining estimations of GT experts' uncertainty and classification ensemble allowed us to achieve the best results (\textbf{2.5}-fold enhancement compared to CE).
Its approximations without using prior knowledge achieve significant enhancement in comparison to CE: in area above curve  combination of $\mathrm{EAE_{MP}}$ and $\mathrm{EXP_{STD}}$ is \textbf{1.35} times better; $\mathrm{EAE_{MP}}$ provides \textbf{0.22\%} positive delta in accuracy when discarding 10\% of data; combination of $\mathrm{EAE_{MP}}$ and $\mathrm{CE_{STD}}$ is \textbf{2.36} times better in terms of reaching 99\% accuracy. 
It is also worth mentioning that a combination of $\mathrm{EAN_{MP}}$ and $\mathrm{CE_{STD}}$ works almost as well as other proposed methods while requires to fine-tune only one net instead of an ensemble.

\section{Conclusions}

Our study introduces an approach to enhance uncertainty estimation in neural networks for medical decision-making by incorporating expert assessments alongside traditional target labels. 
This method, tested in blood typing with a unique dataset featuring complexity scores from medical specialists, significantly improves uncertainty prediction accuracy. 
We observed a 2.5-fold increase in the quality of uncertainty estimation using expert labels and a 35\% increase with neural-based expert complexity estimates compared to existing models. 
This dual-input strategy would help to define best practices for developing more reliable and interpretable neural network models in healthcare and beyond.


\bibliographystyle{splncs04}

\bibliography{article_bib}

\begin{thebibliography}{10}
\providecommand{\url}[1]{\texttt{#1}}
\providecommand{\urlprefix}{URL }
\providecommand{\doi}[1]{https://doi.org/#1}

\bibitem{abdullah2022review}
Abdullah, A.A., Hassan, M.M., Mustafa, Y.T.: A review on bayesian deep learning in healthcare: Applications and challenges. IEEE Access  \textbf{10},  36538--36562 (2022)

\bibitem{ayhan2020expert}
Ayhan, M.S., K{\"u}hlewein, L., Aliyeva, G., Inhoffen, W., Ziemssen, F., Berens, P.: Expert-validated estimation of diagnostic uncertainty for deep neural networks in diabetic retinopathy detection. Medical image analysis  \textbf{64},  101724 (2020)

\bibitem{bishop2006pattern}
Bishop, C.M.: Pattern recognition and machine learning. Springer google scholar  \textbf{2},  5--43 (2006)

\bibitem{briganti2020artificial}
Briganti, G., Le~Moine, O.: Artificial intelligence in medicine: today and tomorrow. Frontiers in medicine  \textbf{7}, ~27 (2020)

\bibitem{campos2023evolving}
de~Campos~Souza, P.V., Lughofer, E.: Efnc-exp: An evolving fuzzy neural classifier integrating expert rules and uncertainty. Fuzzy Sets and Systems  \textbf{466},  108438 (2023)

\bibitem{denker1990transforming}
Denker, J., LeCun, Y.: Transforming neural-net output levels to probability distributions. NeurIPS  \textbf{3},  853--859 (1990)

\bibitem{eng2021artificial}
Eng, D.K., Khandwala, N.B., Long, J., Fefferman, N.R., Lala, S.V., Strubel, N.A., Milla, S.S., Filice, R.W., Sharp, S.E., Towbin, A.J., et~al.: Artificial intelligence algorithm improves radiologist performance in skeletal age assessment: a prospective multicenter randomized controlled trial. Radiology  \textbf{301}(3),  692--699 (2021)

\bibitem{gal2016dropout}
Gal, Y., Ghahramani, Z.: Dropout as a bayesian approximation: Representing model uncertainty in deep learning. In: ICML. pp. 1050--1059 (2016)

\bibitem{gawlikowski2023survey}
Gawlikowski, J., Tassi, C.R.N., Ali, M., Lee, J., Humt, M., Feng, J., Kruspe, A., Triebel, R., Jung, P., Roscher, R., et~al.: A survey of uncertainty in deep neural networks. Artificial Intelligence Review  \textbf{56}(1),  1513--1589 (2023)

\bibitem{gruber2023sources}
Gruber, C., Schenk, P.O., Schierholz, M., Kreuter, F., Kauermann, G.: Sources of uncertainty in machine learning--a statisticians' view. arXiv preprint  (2023)

\bibitem{herron2020ensembles}
Herron, E.J., Young, S.R., Potok, T.E.: Ensembles of networks produced from neural architecture search. In: IEEE HiPC. pp. 223--234 (2020)

\bibitem{hinton1993keeping}
Hinton, G.E., Van~Camp, D.: Keeping the neural networks simple by minimizing the description length of the weights. In: COLT. pp. 5--13 (1993)

\bibitem{huang2023review}
Huang, L., Ruan, S., Xing, Y., Feng, M.: A review of uncertainty quantification in medical image analysis: probabilistic and non-probabilistic methods. arXiv preprint  (2023)

\bibitem{kabir2018neural}
Kabir, H.D., Khosravi, A., Hosen, M.A., Nahavandi, S.: Neural network-based uncertainty quantification: A survey of methodologies and applications. IEEE Access  \textbf{6},  36218--36234 (2018)

\bibitem{korchagin2024image}
Korchagin, S., Zaychenkova, E., Ershov, E., Pishchev, P., Vengerov, Y.: Image-based second opinion for blood typing. Health Information Science and Systems  \textbf{12}(1), ~28 (2024)

\bibitem{kostenok2023uncertainty}
Kostenok, E., Cherniavskii, D., Zaytsev, A.: Uncertainty estimation of transformers' predictions via topological analysis of the attention matrices. arXiv preprint  (2023)

\bibitem{kurz2022uncertainty}
Kurz, A., Hauser, K., Mehrtens, H.A., Krieghoff-Henning, E., Hekler, A., Kather, J.N., Fr{\"o}hling, S., von Kalle, C., Brinker, T.J.: Uncertainty estimation in medical image classification: systematic review. JMIR  \textbf{10}(8),  36427 (2022)

\bibitem{lakshminarayanan2017simple}
Lakshminarayanan, B., Pritzel, A., Blundell, C.: Simple and scalable predictive uncertainty estimation using deep ensembles. NeurIPS  \textbf{30} (2017)

\bibitem{liu2019comparison}
Liu, X., Faes, L., Kale, A.U., Wagner, S.K., Fu, D.J., Bruynseels, A., Mahendiran, T., Moraes, G., Shamdas, M., Kern, C., et~al.: A comparison of deep learning performance against health-care professionals in detecting diseases from medical imaging: a systematic review and meta-analysis. The lancet digital health  \textbf{1}(6),  271--297 (2019)

\bibitem{malinin2018predictive}
Malinin, A., Gales, M.: Predictive uncertainty estimation via prior networks. Advances in neural information processing systems  \textbf{31} (2018)

\bibitem{mukhoti2023deep}
Mukhoti, J., Kirsch, A., van Amersfoort, J., Torr, P.H., Gal, Y.: Deep deterministic uncertainty: A new simple baseline. In: Proceedings of the IEEE/CVF CVPR. pp. 24384--24394 (2023)

\bibitem{punzi2024ai}
Punzi, C., Pellungrini, R., Setzu, M., Giannotti, F., Pedreschi, D.: Ai, meet human: Learning paradigms for hybrid decision making systems. arXiv preprint  (2024)

\bibitem{raghu2019direct}
Raghu, M., Blumer, K., Sayres, R., Obermeyer, Z., Kleinberg, B., Mullainathan, S., Kleinberg, J.: Direct uncertainty prediction for medical second opinions. In: ICML. pp. 5281--5290 (2019)

\bibitem{ramalho2020density}
Ramalho, T., Miranda, M.: Density estimation in representation space to predict model uncertainty. In: Engineering Dependable and Secure Machine Learning Systems: Third International Workshop. pp. 84--96 (2020)

\bibitem{sensoy2018evidential}
Sensoy, M., Kaplan, L., Kandemir, M.: Evidential deep learning to quantify classification uncertainty. Advances in neural information processing systems  \textbf{31} (2018)

\bibitem{welling2011bayesian}
Welling, M., Teh, Y.W.: Bayesian learning via stochastic gradient langevin dynamics. In: ICML. pp. 681--688 (2011)

\bibitem{wilder2020learning}
Wilder, B., Horvitz, E., Kamar, E.: Learning to complement humans. arXiv preprint  (2020)

\bibitem{wilson2020bayesian}
Wilson, A.G., Izmailov, P.: Bayesian deep learning and a probabilistic perspective of generalization. NeurIPS  \textbf{33},  4697--4708 (2020)

\end{thebibliography}

\end{document}